\documentclass[runningheads]{llncs}
\usepackage[T1]{fontenc}
\usepackage{graphicx}
\usepackage[colorlinks=true,linkcolor=blue]{hyperref}
\usepackage{color}

\hypersetup{citecolor=blue}

\usepackage{footmisc}
\usepackage{csquotes}
\usepackage{amsmath,amssymb}
\usepackage{mathtools}
\usepackage[inline]{enumitem}
\usepackage[ruled,linesnumbered]{algorithm2e}
\SetAlFnt{\small}
\usepackage{algorithmic}
\usepackage{setspace}
\usepackage{booktabs}
\usepackage{svg}
\usepackage[misc]{ifsym}
\usepackage[title]{appendix}
\usepackage{array}
\newcolumntype{P}[1]{>{\centering\arraybackslash}p{#1}}
\begin{document}
\title{PBRE: A Rule Extraction Method from Trained Neural Networks Designed for Smart Home Services}
%
%
\author{Mingming Qiu\inst{1,2}\textsuperscript{(\Letter)} \and
Elie Najm\inst{1} \and
Rémi Sharrock\inst{1}\and
Bruno Traverson\inst{2}}
\authorrunning{M. Qiu et al.}
%
\institute{Télécom Paris, Palaiseau, France\\
\email{\{Mingming.Qiu, Elie.Najm, Remi.Sharrock\}@telecom-paris.fr}\\\and
EDF R\&D, Palaiseau, France\\
\email{\{Bruno.Traverson\}@edf.fr}}
\maketitle                 
\begin{abstract}
Designing smart home services is a complex task when multiple services with a large number of sensors and actuators are deployed simultaneously. It may rely on knowledge-based or data-driven approaches. The former can use rule-based methods to design services statically, and the latter can use learning methods to discover inhabitants' preferences dynamically. However, neither of these approaches is entirely satisfactory because rules cannot cover all possible situations that may change, and learning methods may make decisions that are sometimes incomprehensible to the inhabitant. In this paper, PBRE (Pedagogic Based Rule Extractor) is proposed to extract rules from learning methods to realize dynamic rule generation for smart home systems. The expected advantage is that both the explainability of rule-based methods and the dynamicity of learning methods are adopted. We compare PBRE with an existing rule extraction method, and the results show better performance of PBRE. We also apply PBRE to extract rules from a smart home service represented by an NRL (Neural Network-based Reinforcement Learning). The results show that PBRE can help the NRL-simulated service to make understandable suggestions to the inhabitant.
\end{abstract}
\section{Introduction}
\label{sec:introduction}
Numerous smart home applications are rapidly emerging to provide various services to inhabitants. Most of these applications belong to the knowledge-based approaches. Expert systems \cite{jackson1986introduction} are one of the most well-known knowledge-based systems. They allow inhabitants to design their services based on a set of rules. However, despite the potential security and privacy risks \cite{surbatovich2017some},
developing smart home services with knowledge-based approaches is usually a complicated manual process, especially when the services are complex or the actuators are diverse and tightly interconnected. Moreover, it is not easy to design rules when only ultimate objectives are known, e.g., it is cumbersome to design rules for an HVAC (heating, ventilation, and air conditioning) system when the desired indoor temperature is specified along with the energy consumption to be minimized.

Other applications to implement a smart home system mainly belong to data-driven approaches. These approaches make strategic decisions based on the analysis and interpretation of data. And learning methods that learn from data and make predictions based on it are at the forefront of data-driven decision making \cite{brunton2022data}. They try to automatically discover the patterns of the systems by analyzing the datasets provided. Thus, in a smart home, these approaches can figure out regulation solutions by studying inhabitants' activities. It is essential to consider the reactions of an inhabitant when attempting to design a user-friendly smart home system \cite{chan2008review}. Reinforcement learning (RL) \cite{franccois2018introduction}, whose basic idea is that an artificial agent learns the system's behavior patterns by interacting with the environment, can consider the inhabitant's reactions to the proposed actions to find out his habitual behaviors, and a group of habitual behaviors can be translated into a service. In this way, RL enables the inhabitant to participate in the control of smart home services. 
Moreover, neural network-based reinforcement learning (NRL), which integrates neural networks with RL, facilitates the modeling of high-dimensional systems for RL \cite{franccois2018introduction}. However, NRL works like a black box as it does not explain why it proposes new services or modifies existing ones.

To overcome the above shortcomings of the two approaches, we propose to extract rules from a trained NRL. The extracted rules allow showing the inhabitant in which situations the NRL suggested certain actions and enrich the knowledge base, which saves the inhabitant from manually creating rules. 
However, most of the existing rule extraction methods focus either only on neural networks with discrete inputs or on binary classification problems. Nevertheless, in a smart home, there are both discrete (window state: open or closed) and continuous states (light intensity or temperature). Moreover, the control of smart home services is not only a binary but also a multi-class classification problem.
In this paper, a method called PBRE (Pedagogic Based Rule Extractor) is proposed to extract rules from a trained NRL that takes discrete and continuous states as input and proposes states for multiple actuators.

In the rest of the paper, Section \ref{sec:context_and_related_work} presents existing work on rule extraction. Section \ref{sec:pedagogical_rule_extractor} explains the principle of PBRE. Section \ref{sec:compatative_experiment} evaluates PBRE and compares it with an existing method called RxNCM.\footnote[1]{\label{foot:experiment}{The codes for the implementations of all experiments can be found in:\\ } \url{https://github.com/mingming81/PBRE.git}} 
Section \ref{sec:ARL_rule_extraction_smart_home_application} shows how the NRL learning and rule extraction methods can be integrated into a smart home system. 
Section \ref{sec:indoor_ligt_intensity_control_experiment} simulates smart home light services with NRLs and evaluates the performance of PBRE in extracting rules from NRLs.\footref{foot:experiment} Section \ref{sec:conclusion} summarizes the main contributions and provides interesting perspectives.

\section{Context and Related Work}
\label{sec:context_and_related_work}
The smart home is usually implemented by setting up various services. In our study, a set of possible operations performed by different devices can change the value of a particular environment state. A service can be denoted by the name of that state, and the operations involved are means to implement that service. For example, if an inhabitant is at home, raising the heater to increase the temperature or opening a window to decrease the temperature are two ways to implement a temperature service.

To create services, we can use methods of knowledge-based \cite{garcia2014midgar,leong2009rule,mainetti2015novel} or data-driven approaches. 
However, knowledge-based systems usually require manual input from the inhabitant to design services, which hinders the creation of complex services.
Although neural networks are more and more popular, and NRLs are used by many smart home systems \cite{lee2019reinforcement,xu2020multi,yu2019deep} to create services that can interact with the environment and adapt to the inhabitant's activities,
they are like black boxes, and we do not know why they suggest certain services.

To make NRLs understandable and enrich the knowledge base, we consider extracting rules from NRLs. There is a lot of work on extracting rules from trained neural networks. For example, in a tree-based machine learning approach \cite{kern2019tree}, \cite{bride2018towards} collects all formulas from the root to a leaf node with a decision value and conjugates all these formulas to obtain a rule. 
\cite{biswas2017rule} proposes an algorithm called RxNCM. This algorithm first removes insignificant input attributes from the trained neural network. Then, it determines the ranges for each attribute by selecting its minimum and maximum values from the training samples. The rules are created by combining attributes with ranges of values and the corresponding outputs. These rules are then pruned by removing conditions from a rule if the accuracy of the rules can be increased in the test dataset. Finally, the pruned rules are updated by removing overlapping ranges of attribute values between rules if the accuracy of the new rules in the test dataset is increased.  
\cite{towell1993extracting} proposes MOFN to extract rules. First, a neural network is created using KBANN \cite{towell1993symbolic} and then trained. Next, units with similar weighted connections are grouped. The average values replace the weight values in each group, and the groups with low link weights are deleted. The updated neural network is subsequently trained again by optimizing the bias values. Then rules with weights and biases are extracted by combining inputs and outputs. The final rules are obtained by removing the weights and biases. 
However, with the exception of RxNCM, the above work focuses on neural networks that either have specific structures, e.g., \cite{bride2018towards} is only suitable for tree-based machine learning methods, or they only accept categorical and limited integer inputs, e.g., \cite{towell1993extracting} is only suitable for neural networks with limited integer inputs. In this work, we propose PBRE to extract rules from trained neural networks or NRLs by ignoring the input data types and structures of neural networks, and then evaluate it with RxNCM to prove its better performance.

\section{The Proposed PBRE Method}
\label{sec:pedagogical_rule_extractor}

\begin{figure}[t!]
\centering
 \begin{minipage}[t]{0.43\textwidth}
     \centering
     \includegraphics[width=\textwidth]{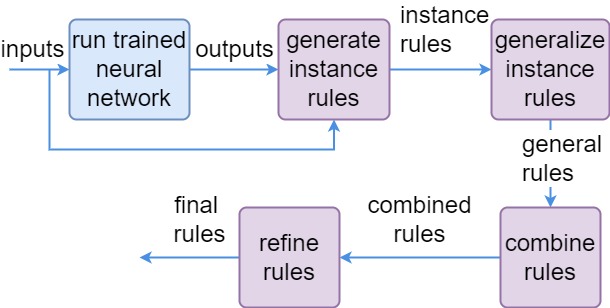}
     \caption{PBRE rule extraction process}
    \label{fig:pedaPrinciple}
   \end{minipage}
   \begin{minipage}[t]{0.561\textwidth}
     \centering
     \includegraphics[width=\textwidth]{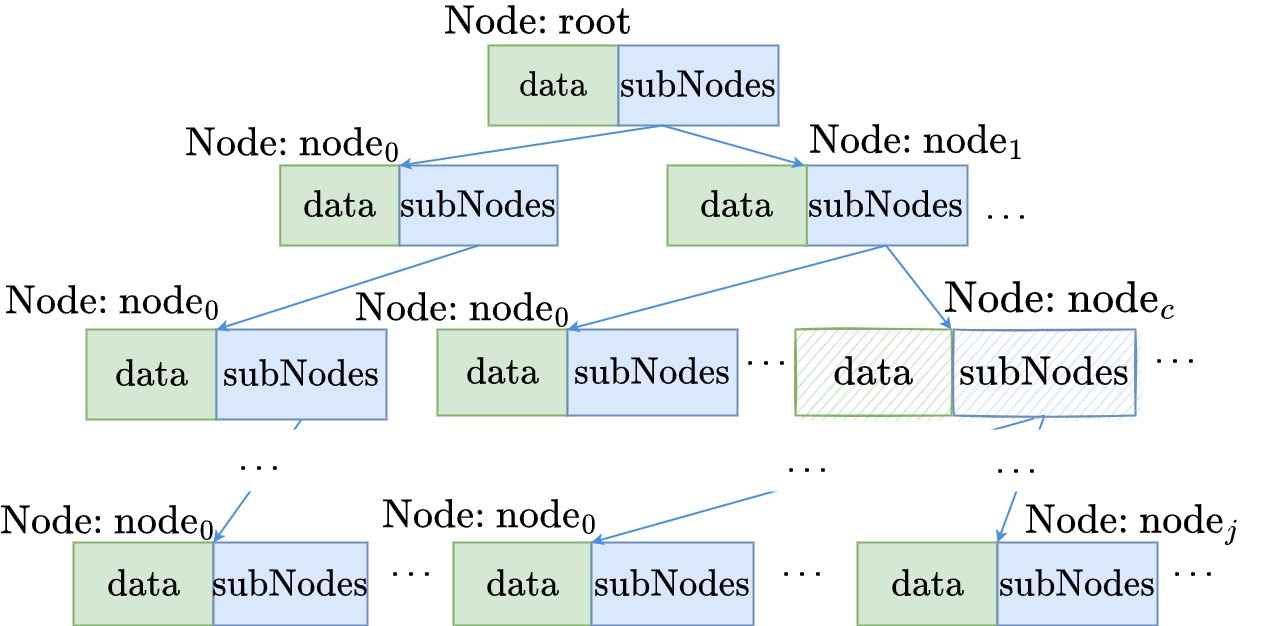}
     \caption{Tree data structure}
  \label{fig:tree_data_structure}
   \end{minipage}
\end{figure}

The principle of PBRE is illustrated in Fig.\ref{fig:pedaPrinciple}: First, PBRE extracts an instance rule from a trained neural network, where an instance rule is a mapping between the inputs and outputs of a neural network at a given time step. Then, it generalizes the instance rule. Next, it combines the generalized rules by merging those whose conclusions are the same and the range values of the states in the conditions overlap. Finally, it refines the combined rules by removing insignificant states in the conditions based on the accuracy of the rules in the unseen dataset. The unseen dataset is a dataset that contains samples from which no rules are extracted, while the seen dataset is used to extract rules. The use of the unseen dataset allows the evaluation of the generalization capability \cite{kamruzzaman2010extraction,taha1999symbolic} of the extracted rules. 

\subsection{Generate Instance Rules}
\label{sec:generate_instance_rule}
A neural network can directly output the desired action that an actuator will perform. Nevertheless, an NRL may provide the quality values for all possible states of an actuator, e.g., a deep neural network-based Q-learning (DQN) \cite{hester2018deep}, which we use in this paper to implement smart home services. Therefore, we first introduce how to select the state of an actuator when the NRL generates quality values for all possible states of that actuator. Suppose that there is a trained NRL with an unknown structure, its inputs are $\mathbf{s}_t=[s^{i^0}_{t},\cdots,s^{i^n}_{t}]$ where the superscript indicates the type of input state and the subscript denotes the time step. The outputs are action quality values $\mathbf{q}_t= [q^{{act}^0}_{0,t},\cdots,q^{{act}^0}_{i,t},\cdots,q^{{act}^m}_{0,t},\cdots,q^{{act}^m}_{k,t}]$, where the superscript indicates the type of actuators and the subscript represents the actuator's certain state and time step. Thus, the proposed states are
\begin{flalign}
{
  \begin{aligned}
  \label{eq:rule_instance_dqn_state}
s^{{act}^0}_{i^{\prime},t}=arg max(q^{{act}^0}_{0,t},\cdots,q^{{act}^0}_{i,t}), \cdots,s^{{act}^m}_{k^{\prime},t}=arg max(q^{{act}^m}_{0,t},\cdots,q^{{act}^m}_{k,t}).
  \end{aligned}
  }
\end{flalign}
Moreover, we can obtain and represent the instance rule ${ir}_t$ at time step $t$ as
\begin{flalign}
{
  \begin{aligned}
  \label{eq:rule_instance_1}
\textit{ if state }i^0 \textit{ has value }s^0_t,\cdots,
\textit{ then }
\textit{actuator }
{act}^0 \textit{ will have state }s^{{act}^0}_{i^{\prime},t}, 
\cdots.
  \end{aligned}
  }
\end{flalign}
However, instance rules are not like rules for general situations. We should generalize them into rules by performing the following procedure.

\subsection{Generalize Instance Rules}
\label{sec:generalize_instance_rules}
We generalize instance rules by first expressing instance rules in a tree structure with linked lists. This tree structure gives a concise and clear idea of an instance rule's compositions and generalization process. Then, we generalize an instance rule by merging it with another instance rule or rule whose conclusions are the same as those of the instance rule and whose certain conditions have close values or contain the values of those of the instance rule.

Fig.\ref{fig:tree_data_structure} shows the structure of the linked list tree. Each branch is a linked list and stores an instance rule or a rule. Each branch node consists of a state value belonging to either conclusions or conditions. Except for the root node whose value is constant, the other nodes at the same level describe the values of the same state type of different instance rules or rules.
The data structure of each node belonging to the $Node$ class is shown in Fig.\ref{fig:tree_data_structure}. It consists of two parts: \textit{data} being the state value of the current node and \textit{subNodes} storing the child nodes of the current node. The state value of the node is represented as
\begin{equation}
\label{eq:node_value}
{
\begin{multlined}
node.data=\{Float:mean,Float:min, Float:max\}
\end{multlined}
}
\end{equation}
where $\textit{mean, min}$, and $\textit{max}$ are the average, minimum, and maximum of the state value of all combined samples. The introduction of a range of values between the minimum and the maximum makes it possible to combine instance rules or an instance rule and a rule when their conclusions are the same, while the values of certain states in the conditions are the same or close or overlap.
The data structure of $subNodes$ is shown in Eq.\ref{eq:subNodes_data_structure}. It is a vector consisting of all subnodes of the current node. Each element in this vector contains a subnode of class $Node$ and an integer $count$ indicating how many times the state value of the current subnode appears simultaneously with the state values of the nodes in the same branch and at higher levels.
\begin{equation}
\label{eq:subNodes_data_structure}
{
\begin{multlined}
\centering
\textit{node.subNodes}=\{\{\textit{Node:{node}}_0,\textit{Integer:{count}}_0\},\cdots
\}.
\end{multlined}
}
\end{equation}

To access a node and its subnodes, we use the dot notation like the property accessor in JavaScript, i.e. the node $n_i$ in slashes in Fig.\ref{fig:tree_data_structure} can be described as $root.{node}_1.{node}_c $, which belongs to the class $Node$, its state value is $n_i.data$, and the subnodes are: $n_i.subNodes$. To access its average, minimum and maximum values, we use $n_i.data.mean, n_i.data.min$ and $n_i.data.max$;
to query its $j^{th}$ subnode, we use $n_i.subNodes[j]$.

However, after generalizing an instance rule, there is a possibility that some particular state value ranges overlap between rules with the same conclusions. Therefore, we need to combine the obtained rules.

\begin{algorithm*}[t!]
\algsetup{linenosize=\tiny}
\setstretch{0.93}
\DontPrintSemicolon 
\SetAlgoLined
\KwResult{final rules}
$D$: dataset storing samples with each containing state values and target values\;
$D_{unseen}$: unseen dataset with each containing state values and target values\;
$R=\{r_1,\cdots,r_m\}$: extracted rules\;
$maxAcc$: the maximum accuracy\;
$stateTypes$: the available state types\;
$insignificantStates$: vector storing insignificant state types\;
\SetKwFunction{FMain}{{\textbf{RIS}}}
  \SetKwProg{Fn}{Function}{:}{}
  \Fn{\FMain{\text{\upshape $D$, $D_{unseen}$, $R$}}}{
  $corrState=D.corr().iloc[:-1,-1]$;\label{algo2:line1}
  sort $corrState$ in ascending order\;\label{algo2:line3}
  sort $stateTypes$ with the same order of $corrState$\;\label{algo2:line4}
  \For{stateType in stateTypes \label{algo2:line11}}{
    $numCorrect=0$\;\label{algo2:line12}
    
    \For{$d_{unseen}$ in $D_{unseen}$ \label{algo2:line13}}{
    
    create $R_1\in R$ without states with types in $insignificantStates$ and $stateType$\;\label{algo2:line13.1}
  
    select $R_2\in R_1$, whose rules' states ranges contain those of $d_{unseen}$ 
    \;\label{algo2:line14.5}
    
    create $R_0$ by adding deleted states to $R_2$  \;\label{algo2:line13.0}
    
    \lIf{$size(R_2)$ \text{equals} $1$}{\label{algo2:line14.6}
        $inference=conclusions$ \text{of }$R_2$\label{algo2:line14.7}
    }
    \Else{ \label{algo2:line14.8}
    
  $arr$=concatenate $d_{unseen}.states$ and $R_0.states.mean$\;\label{algo2:line14}
  
  $arr, corrState2$=remove states with types in $insignificantStates$ and $stateType$ from $arr, corrState$\;\label{algo2:line15}

        $inference=Inference(corrState2,arr)$\label{line:inferenceb}
        }
        \lIf{$inference \text{ equals } d_{unseen}.targets$}{$numCorrect+=1$}
    }
    
    $acc=numCorrect/length(D_{unseen})$\;\label{line:inferencee}
    
    \lIf{$acc\geq maxAcc$ \label{algo2:line22}}{
        add $stateType$ to $insignificantStates$;\label{algo2:line24}
        $maxAcc=acc$\label{algo2:line25}
    }\label{algo2:line26}

  }
  
  \For{stateType in insignificantStates \label{algo2:line28}}{
  \For{$d_{unseen}$ in $D_{unseen}$}{
    $arr$=execute lines \ref{algo2:line13.1}$\sim$\ref{algo2:line14}\;
    $arr,corrState2$=$arr,corrState$ remove states with types in $insignificantStates$ and add state with type $stateType$\;
        execute lines \ref{line:inferenceb}$\sim$\ref{line:inferencee}\;
    }
    \lIf{$acc\geq maxAcc$\label{algo2:line32}}{
        remove $stateType$ from $insignificantStates$;\label{algo2:line34}
        $maxAcc=acc$\label{algo2:line35}
    }
    
  }
  remove states with types in $insignificantStates$ from $R$;\label{algo2:line40}
  {return} R \;\label{algo2:line39}
 }
 \caption{Remove Insignificant States (RIS)}
\label{algo:remove_insigificant_states}
\end{algorithm*}

\subsection{Combine Rules}

To combine rules $R_a$ whose conclusions are the same while the ranges of states in the conditions overlap, for each state $c$ in the condition, we first sort rules of $R_a$ in ascending order of the minimum values of $c$. Next, we name the first rule of $R_a$ as $r_m$. Then we iterate $R_a$. For each rule $r_a$ in $R_a$, its minimum value of state $c$ is compared with the maximum value of the same state in $r_m$. If the former is not larger than the latter, we change $r_m.c.max$ to the larger value between $r_a.c.max$ and $r_m.c.max$. If there is no such rule $r_a$, we store $r_m$ in a new rule set $R_b$, define the current $r_a$ as $r_m$, and compare $r_m$ with the remaining rules of $R_a$. 
After all rules in $R_a$ have been tested, we store the last $r_m$ in $R_b$ to ensure that no $r_m$ is forgotten. Finally, we need to delete duplicate rules from $R_b$ because the above process is performed for each state in the conditions, therefore it may result in duplicate rules. The final $R_b$ stores rules with the same conclusions but without states in the conditions whose ranges overlap with those of other rules.

\begin{algorithm*}[t!]
\algsetup{linenosize=\tiny}
\setstretch{0.93}
\DontPrintSemicolon 
\SetAlgoLined
\KwResult{Inference result}
$corrState$: correlation vector between state types and targets\;
$arr$:matrix concatenating states of the unseen sample and averages of the states in the conditions of the rules\;
$\epsilon$: predefined small number\;
\SetKwFunction{FMain}{{\textbf{Inference}}}
  \SetKwProg{Fn}{Function}{:}{}
  \Fn{\FMain{\text{\upshape $corrState$, $arr$}}}{
  $corr=correlation(arr*corrStates)$\;\label{algo3:line5}
  
  \lIf{$\forall i,j\in corr, (i-j)<\epsilon$ \label{algo3:line6}}{select $conclusions$ of the rule which has the maximum sum of frequencies of occurrence of the conclusions\label{algo3:line7}}
  \lElse{
    select $conclusions$ of the rule with the maximum correlation\label{algo3:line8}
  }
  {return} $conclusions$ \;
 }
 \caption{Infer the unseen sample (Inference)}
\label{algo:getCorrelation}
\end{algorithm*}

\subsection{Refine Rules}
\label{sec:Refine Rules}
In this section, we focus on removing insignificant states in the conditions using Algo.\ref{algo:remove_insigificant_states}. 
First, we use Pearson product-moment correlation coefficients (PPMCC) \cite{benesty2009pearson} to calculates the states-targets correlation vector. This vector stores the correlation between available state types and the targets. Then, we sort this correlation vector in ascending order to ensure the least correlated state types come first (lines \ref{algo2:line1}$\sim$\ref{algo2:line4}).
For each $stateType$ in available state types $stateTypes$, 
first, we define a rule set $R_1$ acquired from $R$ by removing states with types in $insignificantStates$ and $stateType$. Then, we create a rule set $R_2$ containing rules from $R_1$, whose states' range values in the conditions contain the states' values of the unseen sample under study $d_{unseen}$. We also define a rule set $R_0$ equal to $R_2$ with the deleted states added (line \ref{algo2:line11}$\sim$\ref{algo2:line13.0}). If the size of $R_2$ is 1, we select the conclusions of this rule as the targets of $d_{unseen}$ (line \ref{algo2:line14.7}).
Otherwise, we remove states with types in $insignificantStates$ and $stateType$ from the concatenated states matrix which combines the states of $d_{unseen}$ and the averages of the states in the conditions of $R_0$. The states with types in $insignificantStates$ and $stateType$ are also removed from the states-targets correlation vector (lines \ref{algo2:line14.8}$\sim$\ref{algo2:line15}). Next, we use Algo.\ref{algo:getCorrelation} to derive the unseen sample $d_{unseen}$ (line \ref{line:inferenceb}).

As shown in Algo.\ref{algo:getCorrelation}, to obtain inference, we first multiply the updated concatenated matrix about states with the states-targets correlation vector, and then use PPMCC to calculate the correlation between $d_{unseen}$ and rules in $R_2$ based on this weighted state matrix (line \ref{algo3:line5}). If all rules have close correlations with $d_{unseen}$, we choose the conclusions of the rule which has the maximum sum of frequencies of occurrence ($count$ in Eq.\ref{eq:subNodes_data_structure}) of the conclusions; otherwise, we select the conclusions of the rule whose states are most strongly correlated with those of $d_{unseen}$ (lines \ref{algo3:line6}$\sim$\ref{algo3:line8}).

After all unseen samples are derived, we compute the accuracy of $R$ without state with types in $insignificantStates$ and $stateType$ (line \ref{line:inferencee}). If the accuracy is not less than the maximum accuracy, the state with type $stateType$ is not important for the rules to correctly make inference. It will be added to the $insignificantStates$ vector, and the current accuracy will be the maximum accuracy (line \ref{algo2:line26}). Next, after having run through all $stateTypes$ and obtained the final $insignificantStates$, we decide which $stateType$ in $insignificantStates$ can be re-added to rules to maintain or improve the accuracy on the unseen dataset. If such a $stateType$ exist, it will be removed from $insignificantStates$ vector, and the updated accuracy will be the new maximum accuracy as shown in lines \ref{algo2:line28}$\sim$\ref{algo2:line35}. When the updated final $insignificantStates$ is acquired, we remove states belonging to types in $insignificantStates$ from $R$ and return the updated $R$ as the final extracted rules (line \ref{algo2:line39}). 
An extracted rule after having been converted to \enquote{if-then} rule can be written as:
\begin{flalign}
{
  \begin{aligned}
  \label{eq:example_final_extracted_rules}
&\textit{if state }i^0 \textit{ is between }s^0_0 \textit{ and }s^0_1 \textit{ and has average }s^0_m,
\cdots,
\textit{ then }
\textit{actuator }\\
&{act}^0 
\textit{ will have state }s^{{act}^0}_{0} \textit{ with frequency of occurrence } {count}^{{act}^0}_0, 
\cdots.
  \end{aligned}
  }
\end{flalign}

\section{Evaluation and Comparison with Existing Work}
\label{sec:compatative_experiment}
Before applying PBRE to extract rules from DQNs in a smart home, we compare and evaluate the performance of PBRE with that of RxNCM on six datasets from the machine learning repository of the University of California Irvine : the Iris dataset, the Wisconsin Breast Cancer (WBC) dataset, the Sonar dataset, the German Credit dataset, the Ionosphere dataset, and the Heart Disease dataset. Descriptions of the datasets are provided in Table \ref{tab:datasets} (Appendix \ref{sec:appendix_data_set}).

\subsection{Comparative Experiment}
\label{sec:experiment_results_pre_rxrcm}

\subsubsection{Metrics}\label{sec:metric_pre_rxncm}
We use the following metrics to evaluate PBRE and compare it with RxNCM: 
\begin{enumerate*}[label=(\arabic*),font=\footnotesize] 
\item\label{item:eval_1} The number of extracted rules. \cite{craven2014learning,kamruzzaman2010extraction}
\item\label{item:eval_4} Accuracy describes the number of samples where the updated controllable environment states conform to the inhabitant's habitual behaviors as a percentage of the total samples. \cite{arbatli1997rule,towell1993extracting,zhou2004rule}
\item\label{item:eval_2} Similarity, or fidelity \cite{arbatli1997rule,towell1993extracting,zhou2004rule}, is the number of samples where conclusions derived from the rules are the same with propositions proposed by the neural networks as a percentage of the total samples. 
\item\label{item:eval_3} Inference is the number of samples that the extracted rules can derive as a percentage of the total samples.
\end{enumerate*}
Metrics {\footnotesize\ref{item:eval_4}$\sim$\ref{item:eval_3}} are evaluated for both seen and unseen samples. The procedure for determining the metrics is shown in Fig.\ref{fig:metric_acquire_process} (Appendix \ref{sec:appendix_metric_acquire}).

\subsubsection{Results}\label{sec:results_pre_rxncm}

\begin{figure}[t!]
\centering
   \begin{minipage}[t]{\textwidth}
     \centering
     \includegraphics[width=\textwidth]{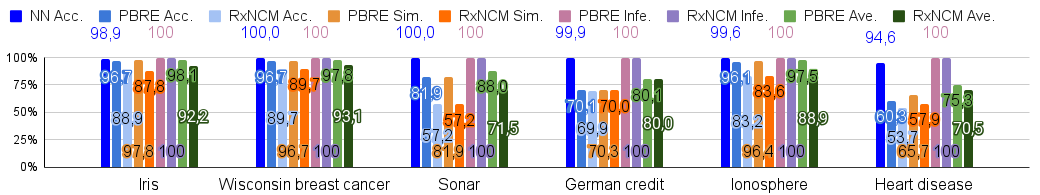}
     \caption{PBRE and RxNCM experiment results by working on seen datasets}
    \label{fig:pre_rxncm_seen}
    \hspace{1em}
   \end{minipage}
   \begin{minipage}[t]{\textwidth}
     \centering
     \includegraphics[width=\textwidth]{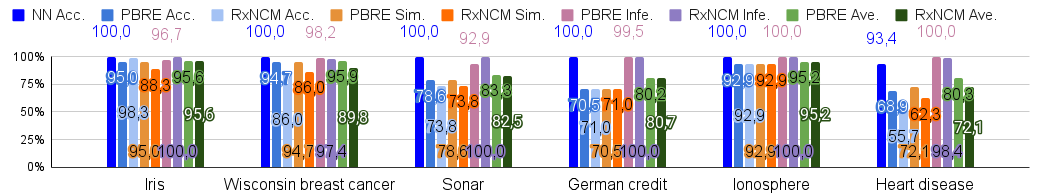}
     \caption{PBRE and RxNCM experiment results by working on unseen datasets}
    \label{fig:pre_rxncm_unseen}
   \end{minipage}
\end{figure}

Table \ref{tab:pre_extract_rules}, Fig.\ref{fig:pre_rxncm_seen} and Fig.\ref{fig:pre_rxncm_unseen} show the metric results for PBRE and RxNCM.
From Table \ref{tab:pre_extract_rules}, we see that the number of rules extracted from each dataset by PBRE or RxNCM is not large, which ensures that storing the extracted rules does not require large memory, which is an important metric for a high-dimensional smart home system. 
Fig.\ref{fig:pre_rxncm_seen} and Fig.\ref{fig:pre_rxncm_unseen} show that although RxNCM, like PBRE, can infer all seen and almost all unseen samples (see \enquote{PBRE Infe.} and \enquote{RxNCM Infe.}), the rules extracted with PBRE generally have higher accuracy and similarity than those extracted with RxNCM (see \enquote{PBRE Acc.}, \enquote{RxNCM Acc.}, \enquote{PBRE Sim.} and \enquote{RxNCM Sim.}). 
To illustrate the general performance, we calculate the average of metrics {\footnotesize\ref{item:eval_4} to \ref{item:eval_3}} and denote it as \enquote{PBRE Ave.} and \enquote{RxNCM Ave.}. 
The results show that PBRE has higher general performance than RxNCM for both datasets, which is consistent with the observations made above for metrics {\footnotesize\ref{item:eval_4} to \ref{item:eval_3}}. 
Moreover, including general performance, RxNCM has higher metric results in the unseen datasets than in the seen datasets, and PBRE does the opposite and has higher metric results than RxNCM in both datasets. This is because to refine rules, PBRE not only deletes the states in the conditions but also adds them back, which ensures the number of states in the conditions and guarantees that PBRE achieves good performance in the unseen datasets and maintains the performance in the seen datasets.

\begin{table*}[t!]
\centering
\caption{Number of rules extracted with PBRE and RxNCM}
\begin{tabular}{c|c|c|c|c|c|c}
\hline
                    & Iris & WBC & Sonar & German credit & Ionosphere & Heart disease \\ \hline
PBRE num. of rules  & 3    & 2                       & 2     & 2             & 2          & 5             \\ \hline
RxNCM num. of rules & 3    & 2                       & 2     & 2             & 2          & 5             \\ \hline
\end{tabular}%
\label{tab:pre_extract_rules}
\end{table*}

\section{NRL and Rule Extraction Methods in the Smart Home}
\label{sec:ARL_rule_extraction_smart_home_application}

\begin{figure}[t!]
\centering
   \begin{minipage}[t]{0.48\textwidth}
     \centering
     \includegraphics[width=\textwidth]{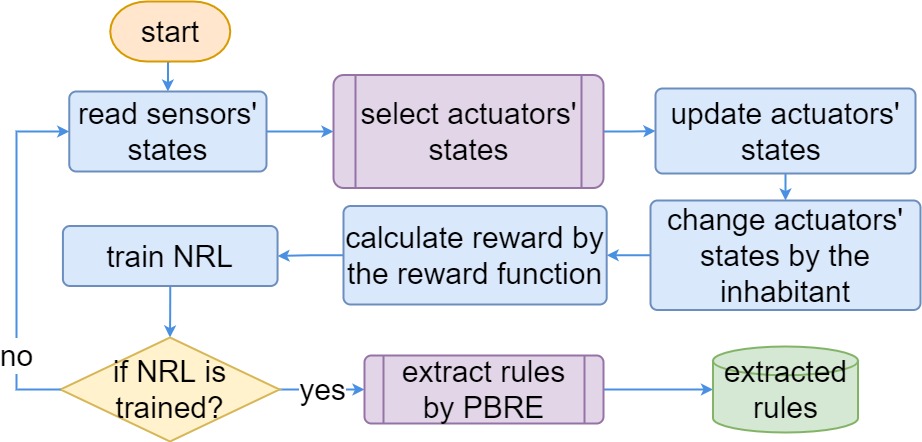}
     \caption{A smart home system with one service in practice }\label{fig:system_working_process_in_reality}
   \end{minipage}\hspace{1em}
   \begin{minipage}[t]{0.48\textwidth}
     \centering
     \includegraphics[width=\textwidth]{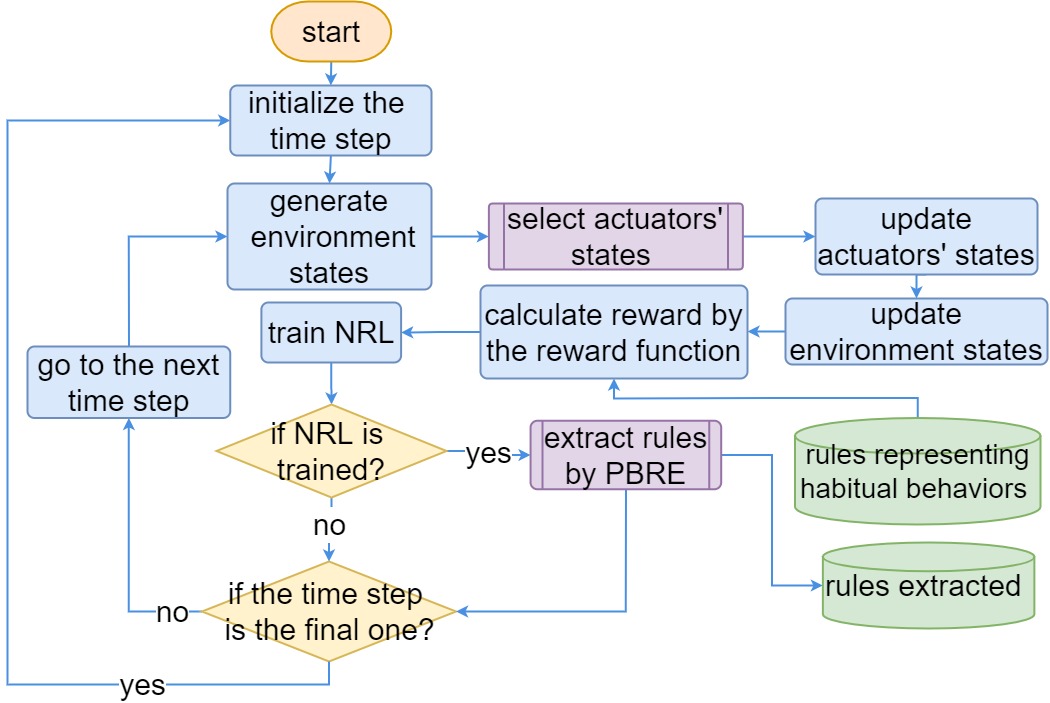}
     \caption{A smart home system with one service in simulation}\label{fig:system_working_process_in_simulation}
   \end{minipage}
\end{figure}

\subsection{Smart Home System in Practice}
\label{sec:ARL_rule_extraction_in_the_reality}

Fig.\ref{fig:system_working_process_in_reality} shows how the smart home system that uses NRL to implement a service and integrates rule extractions looks like in practice: When the system starts, it reads the values of the sensors associated with the service. Then the service selects the involved actuators' states. The actuators update their states to the selected ones. The executions of the actuators lead to changes in the controllable environment states, e.g., indoor light intensity. If the inhabitant is satisfied with the updated controllable states, he takes no action; otherwise, he can change some or all associated actuators' states to match his habitual behaviors. Considering the changes that the inhabitant makes to the actuators' states, the system can then obtain the reward calculated by the predefined reward function. It then trains the NRL using the transitions as input. Each transition contains the environment states detected by the sensors, the states of the actuators selected by the service, the rewards, and the updated environment states. If the NRL is well trained with high and stable accuracy, the system uses PBRE to extract rules and stores them in a database; otherwise, it repeats the above process.

\subsection{Smart Home System in Simulation}
\label{sec:ARL_rule_extraction_in_the_simulation}

Fig.\ref{fig:system_working_process_in_simulation} shows how the smart home system looks like in simulation: First, the time step $t$ is initialized. Next, the environment states, e.g., the inhabitant state, the indoor and outdoor light intensities at time step $t$, are generated by the predefined functions. Then, the service simulated by NRL selects the states that the actuators should take at time step $t$, and the actuators update their states to the selected ones. The controllable environment states are subsequently updated in terms of the actuators' executions. Depending on the predefined reward functions and simulated inhabitant's habitual behaviors, a reward is calculated to indicate whether the inhabitant is satisfied with the updated controllable environment states. After that, the system uses certain optimization algorithm to train the NRL based on the transitions. Once the NRL is well trained with high and stable accuracy, the rules are extracted using PBRE and stored in a database. The system then determines whether the current time step is the last time step. 
However, if the NRL is not well trained, the system directly checks if the current time step is the last one. If not, the system proceeds to the next time step; otherwise, it returns to the first time step and repeats the entire process.

\section{Experiment in the Smart Home Context}
\label{sec:indoor_ligt_intensity_control_experiment}
In this section, we run three tests (DQN\_v1, DQN\_v2, and DQN\_v3), each using a DQN to simulate a smart home light service and evaluating PBRE. To perform these tests, we first introduce a simulated environment and then follow the process in Fig.\ref{fig:system_working_process_in_simulation}, we use PBRE to extract rules from the light service.

\subsection{Simulated Environment}
\label{sec:simulated_experiment}

The representations of the involved variables in the simulated smart home are: 
\begin{enumerate*}[label=(\arabic*),font=\footnotesize] 
\item $s^{us}$: state of the inhabitant;
\item $s^{le}$: outdoor light intensity;
\item $s^{lr}$: indoor light intensity;
\item $s^{lp}$: state of the lamp;
\item $s^{cur}$: state of the curtain.
\end{enumerate*}

Each light service first selects $s^{lp}$ and $s^{cur}$ as its outputs, and only considering $s^{us}$ for DQN\_v1 and also $s^{le}$ for DQN\_v2 and DQN\_v3 as the inputs.
The selected $s^{lp}$ and $s^{cur}$ are used to change $s^{lr}$.
A reward $r$ is then calculated by the predefined reward function with respect to the simulated inhabitant habitual behaviors and the obtained $s^{lr}$.
According to Fig.\ref{fig:system_working_process_in_simulation},
the system trains the DQN by using the transition at each time step. Each transition includes the current environment states $s^{us}$ and $s^{le}$, the proposed actuators states $s^{lp}$ and $s^{cur}$, the reward $r$ and the new environment states $s^{us}$ and $s^{le}$ whose values have not been changed as a result of the actuators' executions. When the DQN is well trained, rules are extracted from it by using the seen dataset as its input, and the simulated smart home system starts a new iteration.
\begin{table*}[t!]
\centering
\def\arraystretch{1.}
\setlength\tabcolsep{6 pt}
\caption{Number of rules extracted by PBRE from different DQNs}
\label{tab:rules_extracted_from_dqns}
\begin{tabular}{ c c c c}
\hline
              & DQN\_v1 & DQN\_v2 & DQN\_v3 \\ \hline
Num. of rules & 4       & 4       & 19      \\ \hline
\end{tabular}%
\end{table*}

\begin{figure}[t!]
   \begin{minipage}[t]{0.49\textwidth}
     \centering
     \includegraphics[width=\textwidth]{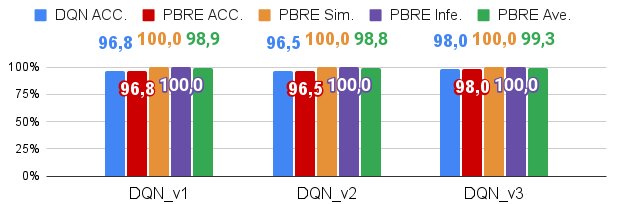}
     \caption{PBRE with the seen datasets}
    \label{fig:nrl_rule_extraction_metrics1}
   \end{minipage}
   \begin{minipage}[t]{0.49\textwidth}
     \centering
     \includegraphics[width=\textwidth]{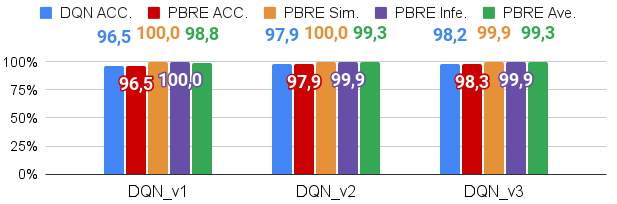}
     \caption{PBRE with the unseen datasets}
    \label{fig:nrl_rule_extraction_metrics2}
   \end{minipage}
\end{figure}
In this experiment, $s^{us}_t$ at time step $t$ is randomly generated by following the uniform distribution ${U}_{int}(0,n_{us})$ which generates an integer between 0 inclusive and $n_{us}$ exclusive,
where $n_{us}$ is the total number of possible states of the inhabitant.
$s^{le}$ within a day is simulated with a Gaussian distribution  \cite{ilyas2012impact,juraimi2009influence}:
\begin{flalign}
{
  \begin{aligned}
  \label{eq:rule_instance_dqn_state}
s^{le}_{t}=\mathcal{N}(amplitude=600,mean=12,stddev=3)+5\cdot U(0,1)
  \end{aligned}
  }
\end{flalign}
where some noise simulated in a uniform distribution $U(0,5)$ with a maximum value of 5 is added. 
$s^{le}_t$ is generated every 5 minutes in one day. 

The output of the light service is $s^{lp}_t$ and $s^{cur}_t$. $s^{lp}_t$ can be selected from multiple levels where level 0 indicates that the lamp is off and other levels represent that the lamp is on. The light intensity that $s^{lp}_{t}$ can provide when it is on is $\beta\cdot s^{lp}_{t}$, where $\beta$ is the light intensity that one level can provide. $s^{cur}_{t}\in \{0,1/2,1\}$ where 0 is closed, 1/2 is half-open, and 1 is fully open.
$s^{lr}_t$ thus is
\begin{equation}
{
    \centering
    \label{eq:indoor_light_intensity}
    \begin{aligned}
    s^{lr}_{{t}}=\beta\times s^{lp}_{{t}}+s^{le}_{{t}}\times s^{cur}_{{t}}.
    \end{aligned}
    }
\end{equation}

The inhabitant's habitual behaviors, which are the ultimate objectives that DQNs try to achieve, are simulated in \enquote{if-then} rules related to $s^{us}$ and $s^{lr}$. For example: \textit{if the inhabitant is absent, then the indoor light intensity is 0 lux}. We do not use specific actuators' states,
which contributes to implementing a smart home service that can derive the regulation solutions when only the ultimate objectives of the inhabitant are given. 
The reward function\footref{foot:experiment} used defines rewards as constant numerical values when different habitual behaviors are satisfied.

\subsection{Experiment Results}
\label{sec:experiment_results}

The metric results of PBRE in extracting rules from different DQNs for the seen and unseen datasets are shown in Table \ref{tab:rules_extracted_from_dqns}, Fig.\ref{fig:nrl_rule_extraction_metrics1} and Fig.\ref{fig:nrl_rule_extraction_metrics2}. 
We can see that the number of extracted rules in Table \ref{tab:rules_extracted_from_dqns} for each simulated service is not large, which ensures that storing these rules does not require large memory. Moreover, we can see that the number of rules in DQN\_v3 has a larger value because the corresponding habitual behaviors of the inhabitant are more complex, as shown in Table \ref{tab:pre_extracted_rules_selected_nrls} (Appendix \ref{sec:appendix_light_rules}). 
Furthermore, from Fig.\ref{fig:nrl_rule_extraction_metrics1} and Fig.\ref{fig:nrl_rule_extraction_metrics2}, we see that PBRE can extract rules from DQNs with satisfactory general performance (see \enquote{PBRE Ave.}), which can be further explained as follows: The extracted rules achieve the same and sometimes even higher accuracy than the DQNs; they have the same similarity to the DQNs in the seen datasets and almost the same similarity in the unseen datasets; moreover, they can infer all seen and almost all unseen samples. One of the extracted rules after having been deleted averages and frequencies of occurrence from Eq.\ref{eq:example_final_extracted_rules} in DQN\_v3 is: 
\textit{if the inhabitant is working, and the outdoor light intensity is between 0.35 and 243.28, then the lamp is at level 3, and the curtain is closed.}
More rules can be found in Table \ref{tab:pre_extracted_rules_selected_nrls} (Appendix \ref{sec:appendix_light_rules}) where we approximate the lowest and highest values of each state's range to integers.

\section{Conclusion}
\label{sec:conclusion}
NRLs can implement smart home services by interacting with the inhabitant and adapting to his habitual behaviors. Yet, like other neural networks, NRLs are black boxes, and the inhabitant cannot know why they suggest certain services.

To address this problem, several contributions are made:
\begin{enumerate*}[label=(\arabic*),font=\footnotesize] 
\item We propose PBRE to extract rules from trained neural networks or NRLs without considering their structures and input data types. And the comparison results with RxNCM prove the better performance of PBRE.
\item We show how the smart home working process, including using an NRL to implement a service and PBRE to extract rules, works in practice and in simulation.
\item We evaluate the performance of PBRE in extracting rules from a smart home light service simulated by a DQN.
The results show that PBRE can satisfactorily extract rules from these DQNs.
\end{enumerate*}

In perspective, it is essential to evaluate the explainability of the extracted rules with qualitative results obtained through focus groups. Then it is promising to work on a proposal for a smart home system with multiple services.

\subsubsection{Acknowledgments}
This work is supported by Seido Laboratory, EDF R\&D Saclay, Télécom Paris, and ANRT (Association Nationale Recherche Technologie) under grant number CIFRE n$^{\circ}$ 2018/1458. 

\subsubsection{Acknowledgments}
This version of the contribution has been accepted for
publication, after peer review but is not the Version of Record and does
not reflect post-acceptance improvements, or any corrections. The Version of Record is
available online at: https://doi.org/10.1007/978-3-031-12426-6\_13

\begin{subappendices}
\renewcommand{\thesection}{\arabic{section}}%

\section{Datasets Descriptions}
\label{sec:appendix_data_set}
The descriptions of the six datasets are in Table \ref{tab:datasets}. 

\section{Metric Acquiring Procedure}
\label{sec:appendix_metric_acquire}
\begin{figure}[b!]
\centering
 \begin{minipage}[t]{0.7\textwidth}
     \centering
     \includegraphics[width=\textwidth]{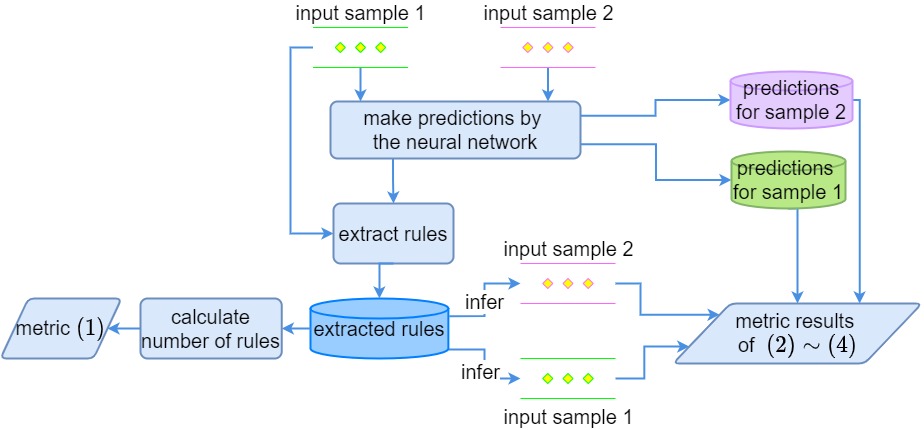}
     \caption{Metrics acquiring procedure}
    \label{fig:metric_acquire_process}
   \end{minipage}\hspace{1em}
\end{figure}

We follow the process in Fig.\ref{fig:metric_acquire_process} to obtain metrics {\footnotesize\ref{item:eval_4}$\sim$\ref{item:eval_3}}. First, we simulate input sample 1 and input sample 2 which are the seen and unseen datasets. The neural network makes predictions and stores them in two databases for the two samples. Next, input sample 1 is used with the neural network to extract rules and obtain metric {\footnotesize\ref{item:eval_1}}.
These rules are used to derive the two samples. Finally, the derived conclusions are compared with the predictions to evaluate metrics {\footnotesize\ref{item:eval_4}$\sim$\ref{item:eval_3}}.

\section{Extracted Rules for Light Services}
\label{sec:appendix_light_rules}

Table \ref{tab:pre_extracted_rules_selected_nrls} shows that DQN\_v1 and DQN\_v2 have the same extracted rules suggesting always closing the curtain. However, the rules in DQN\_v3 have a curtain setting more variable as the energy saving is required in the inhabitant's habitual behaviors. 
Moreover, when expressing the rules, we only keep the states' ranges and forget their averages to make it easier to compare with habitual behaviors.

\begin{table*}[b!]
\centering
\caption{Datasets used for evaluating the performance of the extracted rules}
\def\arraystretch{1.1}
\setlength\tabcolsep{2 pt}
\resizebox{\textwidth}{!}{%
\begin{tabular}{l  P{2cm}  P{3cm} P{4cm} P{3cm}}
\toprule
Dataset                      & Num. of samples & Num. of attributes & Attribute characteristics  & Num. of class  \\ \midrule
Iris dataset                 & 150         & 4           & Real                       & 3    \\
Wisconsin breast cancer      & 569         & 30          & Real                       & 2   \\
Sonar dataset                & 208         & 60          & Real                       & 2    \\
German credit dataset        & 1000        & 9           & Categorical, Integer       & 2    \\
Ionosphere dataset           & 350         & 34          & Integer, Real              & 2   \\
Heart disease dataset        & 303         & 13          & Categorical, Integer, Real & 5    \\
\bottomrule
\end{tabular}
}
\label{tab:datasets}
\end{table*}

\begin{table*}[t!]
\centering
\setlength\tabcolsep{3 pt}
\caption{Extracted Rules for the light service simulated by different DQNs}
\resizebox{\textwidth}{!}{%
\begin{tabular}{  l  p{6.6cm}  p{8.29cm} }\toprule
DQN    & Habitual behaviors & Extracted rules \\\midrule 
v1      & \begin{enumerate*}[label=(\arabic*),noitemsep]
\item If the inhabitant is absent, then the indoor light intensity is 0 lux;
\item If the inhabitant is working, then the indoor light intensity is between 250 lux and 350 lux;
\item If the inhabitant is seeing a movie, then the indoor light intensity is between 350 lux and 450 lux;
\item If the inhabitant is sleeping, then the indoor light intensity is 0 lux.
\end{enumerate*}
& 
\begin{enumerate*}[label=(\arabic*),noitemsep]
\item If the inhabitant is absent, then the lamp is off, and the curtain is closed;
\item If the inhabitant is working, then the lamp is at level 3, and the curtain is closed;
\item If the inhabitant is seeing a movie, then the lamp is at level 4, and the curtain is closed;
\item If the inhabitant is sleeping, then the lamp is off, and the curtain is closed.
\end{enumerate*}
\\\midrule
v2 &
\begin{enumerate*}[label=(\arabic*),noitemsep]
\item If the inhabitant is absent, then the indoor light intensity is 0 lux;
\item If the inhabitant is working, then the indoor light intensity is between 250 lux and 350 lux;
\item If the inhabitant is seeing a movie, then the indoor light intensity is between 350 lux and 450 lux;
\item If the inhabitant is sleeping, then the indoor light intensity is 0 lux.
\end{enumerate*} &
\begin{enumerate*}[label=(\arabic*),noitemsep]
\item If the inhabitant is absent, and the outdoor light intensity is between 0 and 605 lux, then the the lamp is off, and the curtain is closed;
\item If the inhabitant is working, and the outdoor light intensity is between 0 and 605 lux, then the lamp is at level 3, and the curtain is closed;
\item If the inhabitant is seeing a movie, and the outdoor light intensity is between 0 and 605 lux, then the lamp is at level 4, and the curtain is closed;
\item If the inhabitant is sleeping, and the outdoor light intensity is between 0 and 605 lux, then the lamp is off, and the curtain is closed.
\end{enumerate*}
\\\midrule
v3 &
With the preference of decreasing the electricity consumption:
\begin{enumerate*}[label=(\arabic*),noitemsep]
\item If the inhabitant is absent, then the indoor light intensity is 0 lux;
\item If the inhabitant is working, then the indoor light intensity is between 250 lux and 350 lux;
\item If the inhabitant is seeing a movie, then the indoor light intensity is between 350 lux and 450 lux;
\item If the inhabitant is sleeping, then the indoor light intensity is 0 lux.
\end{enumerate*} &
\begin{enumerate*}[label=(\arabic*),noitemsep]
\item If the inhabitant is absent, and the outdoor light intensity is between 0 and 605 lux, then the lamp is off, and the curtain is closed;
\item If the inhabitant is working, and the outdoor light intensity is between 246 and 357 lux, then the lamp is off, and the curtain is fully open;
\item If the inhabitant is working, and the outdoor light intensity is between 512 and 605 lux, then the lamp is off, and the curtain is half-open;
\item If the inhabitant is seeing a movie, and the outdoor light intensity is between 356 and 452 lux, then the lamp off, and the curtain is fully open;
\item If the inhabitant is sleeping, and the outdoor light intensity is between 0 and 605 lux, then the lamp is off, and the curtain is closed; $\cdots$
\end{enumerate*}
\\\bottomrule 
\end{tabular}
}
\label{tab:pre_extracted_rules_selected_nrls}
\end{table*}

\end{subappendices}
%
%
%
\bibliographystyle{splncs04}
\bibliography{mybibliography}
%




\end{document}